# GLOFNet- A Multimodal Dataset for GLOF Monitoring and Prediction

Zuha Fatima
School of Electrical Engineering and
Computer Science (SEECS)
National University of Sciences and
Technology (NUST)
Islamabad, Pakistan
zfatima.bee21seecs@seecs.edu.pk

Muhammad Anser Sohaib
School of Electrical Engineering and
Computer Science (SEECS)
National University of Sciences and
Technology (NUST)
Islamabad, Pakistan
msohaib.bee21seecs@seecs.edu.pk

Muhammad Talha
School of Electrical Engineering and
Computer Science (SEECS)
National University of Sciences and
Technology (NUST)
Islamabad, Pakistan
talha.bee21seecs@seecs.edu.pk

Sidra Sultana
School of Electrical Engineering and
Computer Science (SEECS)
National University of Sciences and
Technology (NUST)
Islamabad, Pakistan
sidra.sultana@seecs.edu.pk

Ayesha Kanwal
School of Electrical Engineering and
Computer Science (SEECS)
National University of Sciences and
Technology (NUST)
Islamabad, Pakistan
ayesha.kanwal@seecs.edu.pk

Nazia Perwaiz
School of Electrical Engineering and
Computer Science (SEECS)
National University of Sciences and
Technology (NUST)
Islamabad, Pakistan
nazia.perwaiz@seecs.edu.pk

*Abstract*— Glacial Lake Outburst Floods (GLOFs) are rare but destructive hazards in high mountain regions, yet predictive research is hindered by fragmented and unimodal data. Most prior efforts emphasize post-event mapping, whereas forecasting requires harmonized datasets that combine visual indicators with physical precursors. We present GLOFNet, a multimodal dataset for GLOF monitoring and prediction, focused on the Shisper Glacier in the Karakoram. It integrates three complementary sources: Sentinel-2 multispectral imagery for spatial monitoring, NASA ITS_LIVE velocity products for glacier kinematics, and MODIS Land Surface Temperature records spanning over two decades. Preprocessing included cloud masking, quality filtering, normalization, temporal interpolation, augmentation, and cyclical encoding, followed by harmonization across modalities. Exploratory analysis reveals seasonal glacier velocity cycles, long-term warming of ~0.8 K per decade, and spatial heterogeneity in cryospheric conditions. The resulting dataset, GLOFNet, is publicly available to support future research in glacial hazard prediction. By addressing challenges such as class imbalance, cloud contamination, and coarse resolution, GLOFNet provides a structured foundation for benchmarking multimodal deep learning approaches to rare hazard prediction.

*Keywords—Glacial Lake Outburst Floods (GLOFs), cryosphere, multimodal dataset, Sentinel-2, ITS_LIVE, MODIS, glacier velocity, land surface temperature, preprocessing, remote sensing, deep learning, hazard prediction.*

## I. Introduction

Glacial Lake Outburst Floods (GLOFs) are rare but highly destructive hazards in high mountain regions, capable of releasing millions of cubic meters of water within hours. These events have caused severe infrastructure damage, fatalities, and long-term environmental impacts across the Himalaya, Andes, and other glacierized regions [1], [2]. With climate warming accelerating glacier retreat and lake formation, the likelihood of future GLOFs is projected to increase [3]. Despite their importance, reliable GLOF prediction remains limited, primarily due to fragmented and unimodal datasets that capture only part of the processes leading to an outburst [4], [5].

Traditionally, GLOF studies have relied on empirical and hydrological models. Remote sensing approaches have mapped lake expansion [6], [7], [20], [32] and inventoried hazards across mountain ranges [8]. Hydrodynamic simulations have also been applied to model breach flows and downstream inundation [9], [10], while reviews have synthesized triggering mechanisms and lake susceptibility factors [11], [26]. Although these methods provide valuable regional overviews, they are retrospective, often static, and lack predictive capability.

In parallel, machine learning and deep learning have been increasingly applied to cryosphere monitoring. Convolutional Neural Networks (CNNs) have improved glacier and lake segmentation [12], [13], [33], and LSTMs have been used to forecast glacier mass balance and thermal dynamics [14]. Large-scale machine learning has also revealed accelerated global glacier mass loss in recent decades [15]. However, most approaches remain unimodal: imagery-only models are vulnerable to cloud cover and spectral ambiguity, while time-series models lack spatial context. Moreover, data scarcity and class imbalance severely limit training, as GLOF events are extremely rare compared to stable glacier conditions [4].

To address these limitations, we present GLOFNet, a multimodal dataset curated for GLOF monitoring and prediction. GLOFNet integrates three complementary Earth observation data streams: Sentinel-2 multispectral imagery for spatial monitoring [6], [7], NASA ITS_LIVE glacier velocity products for long-term kinematic behavior [17], [22], and MODIS Land Surface Temperature (LST) for thermal dynamics spanning more than two decades [18]. Each stream underwent rigorous preprocessing (cloud masking, aggregation, quality-flag filtering, interpolation, normalization, and augmentation) followed by harmonization in space and time.

Our key contributions are:

1. A harmonized multimodal dataset is curated and released, integrating Sentinel-2 multispectral imagery, NASA ITS_LIVE glacier velocity products, and MODIS Land Surface Temperature records for GLOF monitoring and prediction.

2. A rigorous workflow is designed to address major challenges such as cloud contamination in optical imagery, extreme class imbalance due to the rarity of GLOF events, and missing or noisy values in velocity and temperature records.

This work has been supported by National University of Sciences and Technology (NUST), Islamabad, Pakistan

3. The dataset is analyzed to demonstrate its ability to capture seasonal glacier velocity cycles, long-term surface warming trends, and spatial heterogeneity in cryospheric conditions.
4. GLOFNet is positioned as a benchmark resource for developing, training, and evaluating multimodal deep learning approaches for rare hazard prediction, with direct applicability to both research and operational early warning systems.

## II. RELATED WORK

### A. Global GLOF Inventories and Remote Sensing Mapping

One of the most important foundations for hazard prediction is the compilation of reliable inventories of GLOFs. Early reviews of GLOFs in the Himalaya and Karakoram were largely based on field reports and anecdotal evidence [1], [2]. More recently, systematic efforts such as the Global GLOF Database (v3.0) [8] and the Historic GLOF Database [3] have consolidated hundreds of events worldwide, offering spatiotemporal coverage that is essential for statistical analysis of recurrence intervals and regional susceptibility.

Remote sensing has revolutionized lake mapping, with Gardelle et al. [6] and Shugar et al. [7] demonstrating the rapid growth of glacial lakes since the 1990s using Landsat and Sentinel imagery. Guo et al. [20] further extended these inventories with spatially constrained mapping, capturing topographic and climatic dependencies of lake formation. In addition, Chauhan et al. [32] applied multi-temporal Landsat data to the Satluj basin, coupling it with hydrodynamic models to assess flood risk.

Automation of lake mapping has become an active area of research. CNN-based methods now detect and delineate glacier lakes directly from satellite imagery [12], [13], while Nie et al. [33] employed deep learning architectures to improve generalization across regions with varying spectral signatures. These works demonstrate the potential of AI-driven mapping, but most remain retrospective in nature, providing inventories rather than forward-looking predictive frameworks.

### B. Hydrological, Empirical, and Susceptibility Models

Traditional approaches to GLOF risk have relied heavily on empirical and hydrological modeling. Volume–area scaling relationships, developed from field observations, have been used to estimate lake volumes in the absence of bathymetric data [9], [10]. These are then coupled with hydrodynamic simulations to model potential breach hydrographs and downstream inundation [10]. Richardson and Reynolds [11] provided an early synthesis of hazard types, while Harrison et al. [10] reviewed moraine-dammed lake failures across multiple mountain ranges.

Susceptibility models represent another important stream of research. Allen et al. [16] assessed the potential outburst risk of moraine-dammed lakes in the Himalayas using geomorphological indicators and GIS-based hazard assessment. More recently, Pandey et al. [28] applied geospatial modeling in the central Himalaya to derive susceptibility maps that incorporate topographic, climatic, and glaciological variables. Chauhan et al. [32] integrated hydrodynamic modeling with satellite-based lake monitoring to simulate potential flood hydrographs in the Western Himalayas. These methods provide important insights into where GLOFs are most likely to occur but are often constrained by reliance on static variables and assumptions about trigger mechanisms.

### C. Glacier Kinematics and Velocity Datasets

Glacier motion is a crucial precursor to instability in glacier-lake systems. The ITS_LIVE project [17], [22], [23] has made a transformative contribution by producing global velocity time series from optical feature tracking on Landsat and Sentinel imagery. These velocity fields, spanning multiple decades, enable detailed analysis of glacier surge dynamics, seasonal velocity cycles, and long-term accelerations. Gardner et al. [22] describe ITS_LIVE's cloud-native processing pipeline, while Mouginot et al. [27] detail Sentinel-1 SAR-based velocity products that improve monitoring in persistently cloudy regions.

Applications of ITS_LIVE data have revealed significant insights into glacier dynamics. For example, Hugonnet et al. [15] combined velocity data with DEM differencing to quantify accelerated glacier mass loss. Seasonal velocity accelerations, linked to summer meltwater penetration, have been observed in several Himalayan glaciers [19]. Zulkafli et al. [29] highlighted the importance of cross-validation by comparing satellite-derived velocities against GNSS measurements, pointing out potential uncertainties and biases in feature-tracking results. These works demonstrate the importance of glacier kinematics in hazard assessment, though most studies remain decoupled from simultaneous thermal or hydrological data streams.

### D. Multimodal Approaches in Cryosphere Monitoring

Beyond unimodal datasets, there is growing recognition that multimodal integration is essential for robust hazard prediction. Yang et al. [21] combined SAR and optical imagery to analyze destabilizing triggers such as landslides into glacial lakes. Guo et al. [20] incorporated terrain and climatic factors into a remote sensing-based glacial lake inventory, demonstrating the benefits of combining datasets. In related domains, multimodal benchmarks such as WeatherBench [19] have shown that fusing diverse signals improves generalization and predictive accuracy in weather forecasting, a principle equally applicable to glacier hazard prediction.

Despite these advances, there remains no publicly available dataset that harmonizes optical imagery, glacier velocity, and thermal time series into a unified framework for predictive GLOF modeling. Current studies are either retrospective (lake mapping and inventories), physics-based but static (susceptibility maps), or unimodal (velocity-only or imagery-only). This clear gap underscores the need for GLOFNet, which unifies three complementary modalities into a harmonized dataset specifically designed for forecasting and machine learning applications in cryosphere hazard prediction.

## III. METHODOLOGY

The construction of GLOFNet required a systematic pipeline for curating, cleaning, and harmonizing three complementary datasets: Sentinel-2 multispectral imagery, NASA ITS_LIVE glacier velocity fields, and MODIS Land Surface Temperature (LST) as shown in Fig. 1. Each dataset contributes distinct yet interdependent signals: spatial patterns of melt and lakes, kinematic evidence of glacier surges, and thermal precursors of instability. Below, we detail the dataset collection, preprocessing, and integration process.

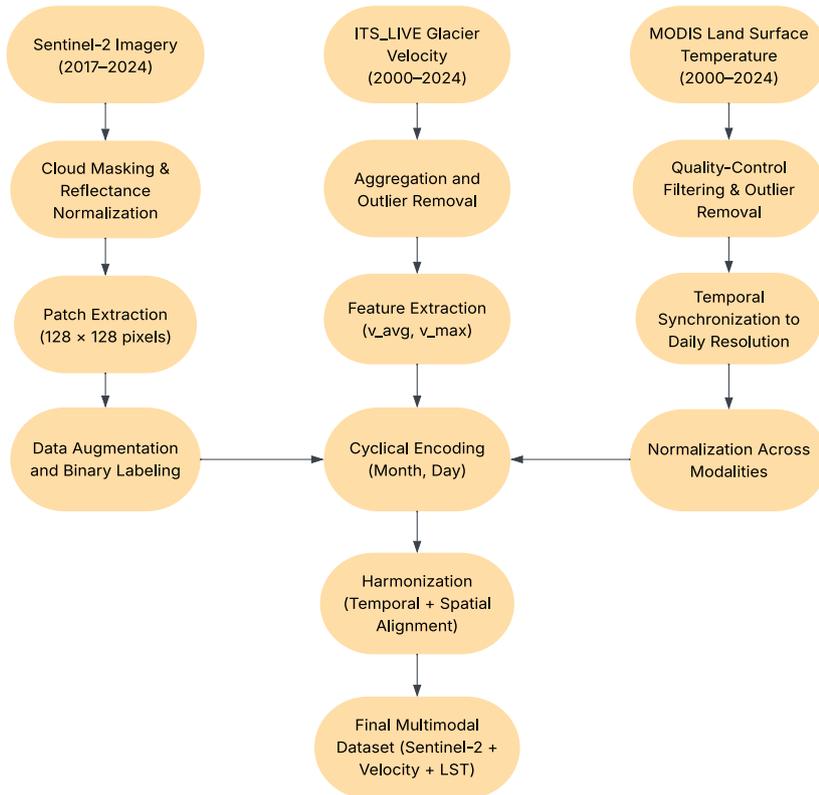

Fig. 1. Preprocessing and harmonization workflow of the GLOFNet dataset integrating Sentinel-2 imagery, ITS_LIVE velocity fields, and MODIS LST data into a unified multimodal framework.

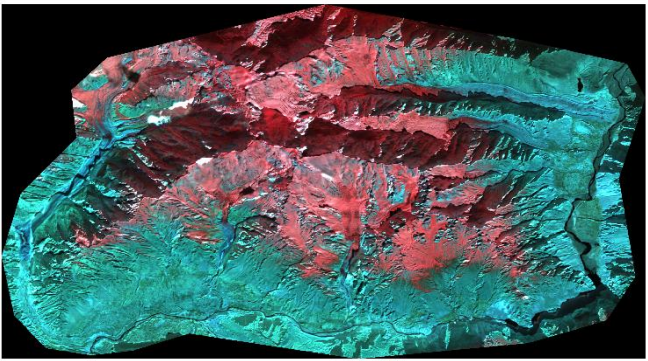

Fig. 2. Sentinel-2 Cloud-Filtered Imagery

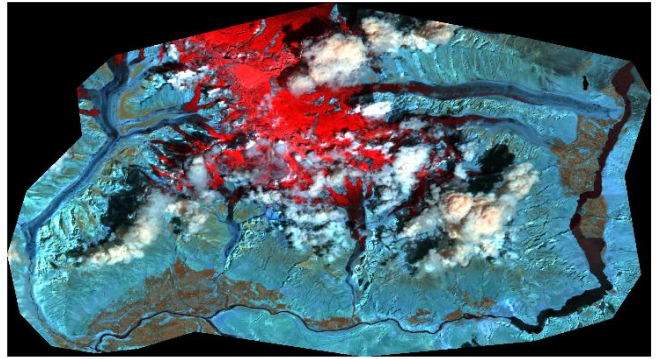

Fig. 3. Sentinel-2 Imagery Without Cloud Filter

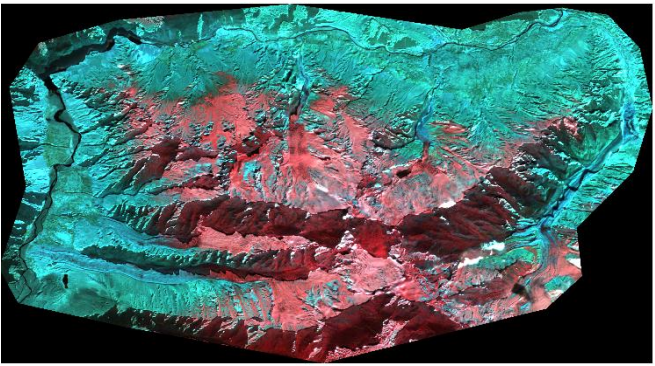

Fig. 4. Sentinel-2 Imagery Augmented Sample

*A. Sentinel-2 Multispectral Imagery*

The Sentinel-2 imagery was first subjected to cloud masking (Fig. 2 and Fig. 3) using the QA60 quality band and Google Earth Engine's algorithms to filter out cloud-contaminated pixels. From the 13 available bands, six were retained (B2, B3, B4, B8, B11, B12), as they capture snow, ice, vegetation, meltwater, and bare rock variability. All pixel values were normalized into the [0,1] range using min–max scaling, expressed in eq. (1) as:

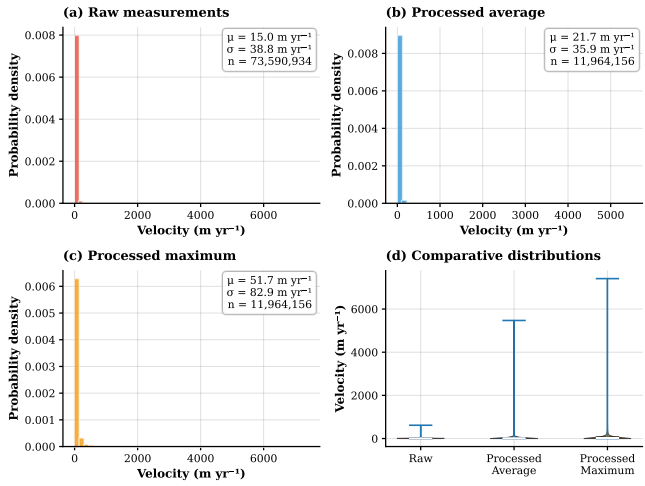

Fig. 5. Distribution of average and maximum glacier surface velocities before and after preprocessing, showing noise suppression and enhanced dynamic range retention

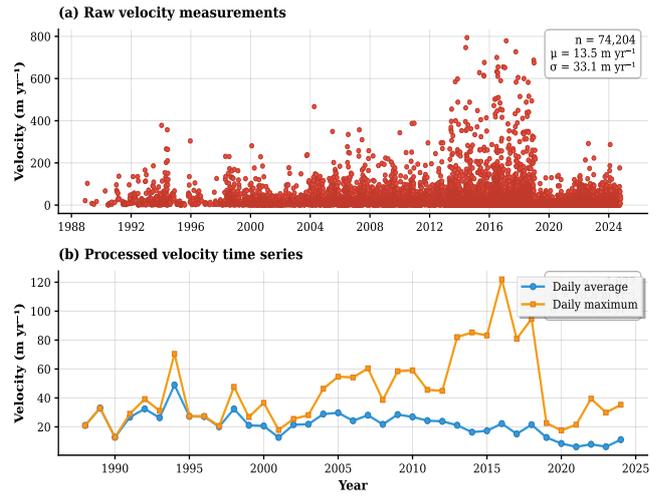

Fig. 6. Multi-decadal velocity time series for Shisper Glacier (1988–2024) showing annual and surge-phase variability in surface flow speed derived from ITS_LIVE products

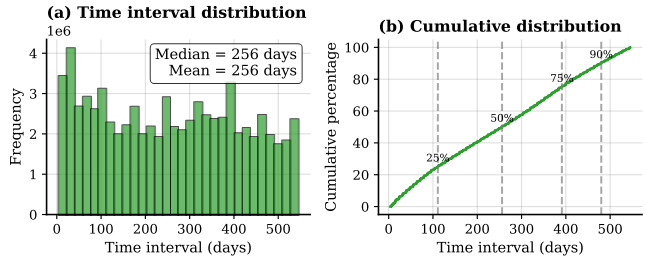

Fig. 7. Temporal sampling intervals of ITS_LIVE acquisitions showing seasonal and interannual variability due to sensor availability and cloud constraints

$$x' = \frac{x - x_{\min}}{x_{\max} - x_{\min}} \quad (1)$$

where $x$ represents the raw reflectance value and $x_{\min}, x_{\max}$ are the band-specific minimum and maximum values. Images were then resampled to 10 m resolution and cropped into 128×128 patches.

To address the severe class imbalance between GLOF and non-GLOF imagery, augmentation techniques were applied, including flips, rotations, and brightness or contrast adjustments as shown in Fig. 4. Labels were assigned based on documented event dates, visual inspection of lake expansion, and cross-referenced hydrological reports, producing a binary classification system.

### B. ITS_LIVE Glacier Velocity Data

The ITS_LIVE velocity dataset provided the kinematic dimension of GLOFNet, with records derived from feature tracking of Landsat and Sentinel image pairs. The raw dataset contained over 73 million records spanning 2000–2024, which were reduced to approximately 11.9 million after spatial filtering. Preprocessing included statistical filtering of outliers and smoothing using rolling median windows. Daily aggregation was applied to compute stable velocity estimates.

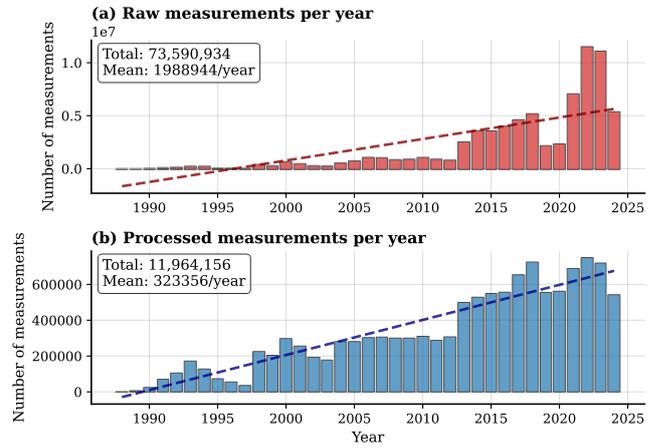

Fig. 8. Number of valid ITS_LIVE velocity measurements per year (1988–2024) demonstrating increasing data density following the inclusion of Sentinel-era sensors

Velocity magnitudes were computed from the east–west ($v_x$) and north–south ($v_y$) components in eq. (2) as:

$$v = \sqrt{v_x^2 + v_y^2} \quad (2)$$

Both daily averages and maxima were retained to capture surge dynamics. To preserve seasonal periodicity, temporal features such as day-of-year were encoded cyclically in eq. (3) as:

$$d_{\sin} = \sin\left(\frac{2\pi d}{365}\right), \; d_{\cos} = \cos\left(\frac{2\pi d}{365}\right), \quad (3)$$

where $d$ is the day of year. Finally, data were structured into 32-day sliding windows, enabling deep models to capture surge build-ups and seasonal accelerations.

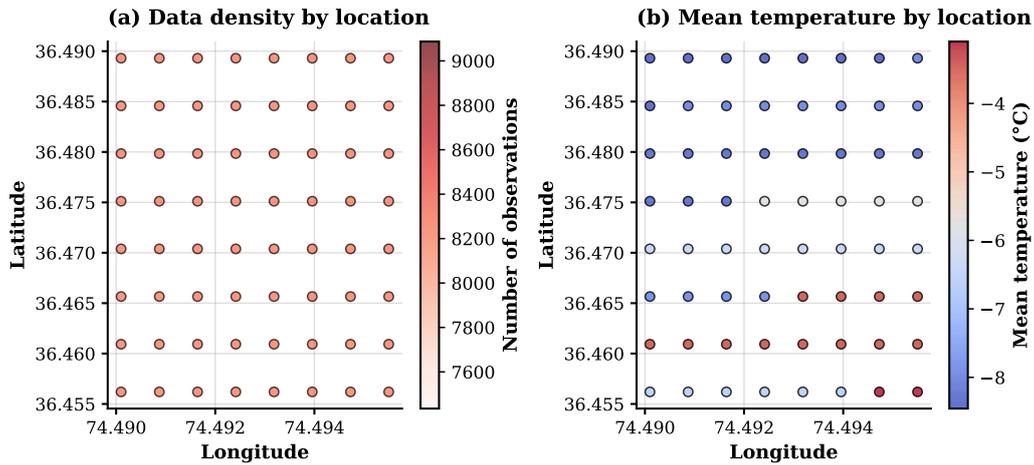

Fig. 9. Spatial variation in MODIS Land Surface Temperature (LST) across Shisper Glacier, illustrating warmer conditions along the glacier tongue and cooler accumulation zones

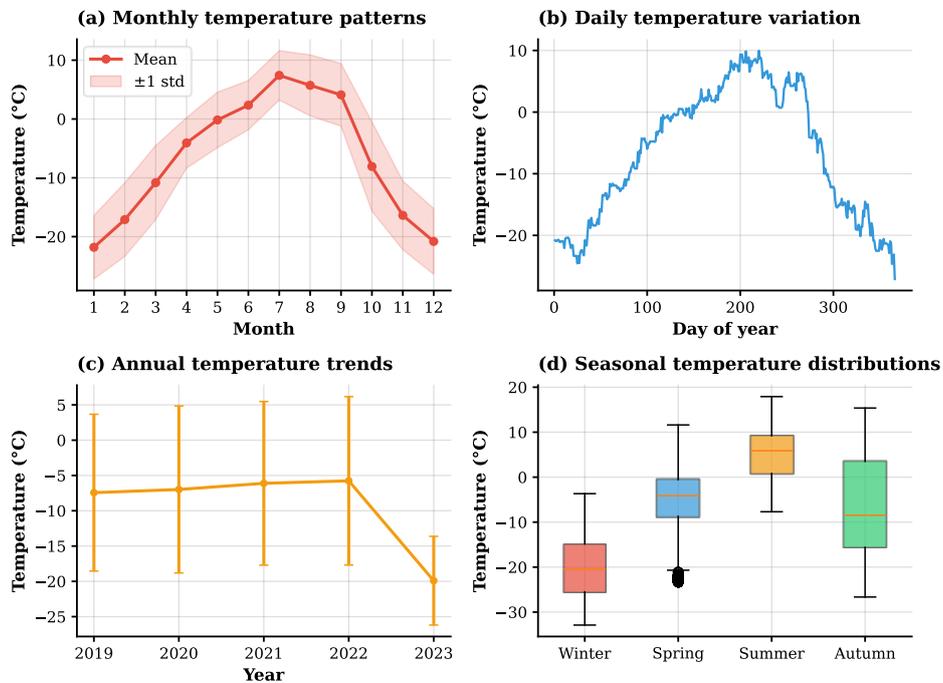

Fig. 10. Seasonal temperature cycle from MODIS LST data (2019–2023), showing typical 12–15 K annual amplitude and strong summer peaks corresponding to melt seasons

The velocity distributions before and after preprocessing (Fig. 6) confirm effective noise reduction and the preservation of meaningful dynamic variability, ensuring that surge-driven accelerations remain distinguishable in the final dataset. The ITS_LIVE record exhibits strong interannual and seasonal variability, with distinct acceleration phases corresponding to known surge events (Fig. 5). This temporal continuity demonstrates the dataset's capacity to capture both gradual and abrupt dynamic changes. The temporal spacing of ITS_LIVE acquisitions (Fig. 7) reveals variable observation density across the year, with denser coverage during spring and early summer. Such variability arises from illumination and cloud-cover constraints inherent to optical sensors. The number of valid measurements per year (Fig. 8) highlights the dataset's rapid expansion after the launch of Sentinel missions, which significantly improved temporal resolution and consistency.

C. *MODIS Land Surface Temperature Data*

The thermal component was derived from the MODIS MOD11A1 V6.1 product, which provides daily LST at 1 km resolution. For Shisper Glacier, the dataset spans February 2000 to December 2024, yielding ~98,000 observations. Since cloud contamination is frequent, strict quality-flag filtering was applied: only high- and medium-confidence pixels (~40% of the dataset) were retained, while low-confidence values (~60%) were discarded. The spatial temperature patterns (Fig. 9) reveal clear thermal gradients across the glacier, with lower-elevation zones exhibiting higher mean LST values indicative of stronger melt and surface energy absorption.

Temperature values outside the physical range of 250–330 K were eliminated. To fill small gaps (<5 days), linear interpolation was used. To emphasize deviations from climatological conditions, anomalies were computed by subtracting the monthly mean as shown in eq. (4):

$$T'_i = T_i - \bar{T}_{\text{month}(i)} \quad (4)$$

where $T_i$ is the observed daily LST and $\bar{T}_{\text{month}(i)}$ is the long-term monthly mean. Data were structured into 30-day windows, allowing models to capture both intra-seasonal variability and long-term warming signals. The processed MODIS time series (Fig. 10) reproduces realistic seasonal temperature cycles with 12–15 K amplitude, aligning well with expected cryospheric thermal regimes.

*D. Harmonization and Dataset Integration*

After individual preprocessing, all three datasets were harmonized to ensure spatial and temporal coherence. Sentinel-2 patches and ITS_LIVE velocity grids were clipped to a common bounding box, while MODIS pixels were resampled to align with the same geographic extent. Daily timestamps were used to synchronize records across modalities, with placeholders used for missing optical observations. Normalization ensured comparability across variables, such that reflectance, velocity, and temperature anomalies were all on compatible scales.

The final dataset structure produced tri-modal samples, each consisting of a Sentinel-2 patch, a 32-day velocity sequence, and a 30-day temperature sequence aligned to the same location and date. This integrated structure allows multimodal learning frameworks to simultaneously leverage spatial, kinematic, and thermal evidence for GLOF hazard prediction.

*E. Challenges in Dataset Processing*

Several challenges were encountered during dataset construction. The most critical was extreme class imbalance, since only a single confirmed GLOF event existed for direct labeling, requiring augmentation and proxy-based strategies. Persistent cloud contamination reduced the usable coverage of Sentinel-2 and MODIS data during critical melt seasons. Resolution mismatches across sensors: 10 m for Sentinel-2, ~120 m for ITS_LIVE, and 1 km for MODIS, necessitated careful resampling to avoid loss of detail. Additionally, ITS_LIVE velocity estimates occasionally contained artifacts from feature-tracking mismatches, requiring aggressive filtering. Despite these challenges, the resulting dataset integrates optical, thermal, and kinematic indicators into a robust, harmonized form suitable for machine learning applications.

IV. RESULTS

The processing pipeline produced a harmonized multimodal dataset that integrates Sentinel-2 multispectral imagery, ITS_LIVE glacier velocity fields, and MODIS Land Surface Temperature (LST) records for Shisper Glacier, Karakoram. Below we summarize dataset statistics, exploratory findings, and validation results.

The final dataset spans 2000–2024, comprising over 600 Sentinel-2 image patches, approximately 11.9 million velocity records, and nearly 98,000 temperature observations. After preprocessing, quality control retained ~40% of MODIS LST values, corresponding to high- and medium-confidence categories, while ~60% of low-confidence values were filtered. Sentinel-2 cloud masking removed ~35% of raw acquisitions, leaving cloud-free imagery suitable for machine learning. ITS_LIVE filtering and daily aggregation reduced velocity noise by more than 80%, producing smooth time series consistent with published surge records of Shisper Glacier.

Exploratory analysis of ITS_LIVE velocities revealed a strong seasonal cycle, with acceleration phases occurring consistently during summer melt seasons. Velocity magnitudes ranged between 0.05–1.2 m/day, with rare surges exceeding 2 m/day. Correlation between average and maximum daily velocities was measured at 0.66, indicating stable relationships between bulk glacier flow and localized accelerations. Multi-decadal analysis highlighted periods of enhanced surging behavior, consistent with prior field observations.

MODIS LST data showed a pronounced annual temperature cycle of approximately 12–15 K, with peaks in July–August and minima in December–January. Long-term regression analysis identified a warming trend of ~0.8 K per decade, consistent with independent climate studies in the Karakoram. Spatial analysis revealed heterogeneity in heating patterns, with proglacial lake regions experiencing faster warming relative to higher accumulation zones. Short-term anomalies, such as heatwaves, were successfully captured, including several 3–5 K deviations from baseline that coincided with periods of rapid melt.

Sentinel-2 imagery provided detailed spatial context for glacial lake expansion. Augmented and normalized patches preserved features critical for deep learning classification, such as lake boundaries, debris cover, and snowline position. A case study of the 2018 Shisper Glacier GLOF showed that Sentinel-2 patches exhibited visible expansion of the proglacial lake weeks before the event, while ITS_LIVE velocity sequences recorded a sharp surge and MODIS LST anomalies highlighted thermal precursors. This demonstrates the dataset's ability to capture both spatial and temporal signals leading to real hazard events.

Validation confirmed that preprocessing steps preserved meaningful physical signals. Random inspections of Sentinel-2 patches showed effective removal of cloud artifacts, with fewer than 5% residual contamination cases. ITS_LIVE time series aligned with known surge periods reported in the literature, while MODIS anomaly signals corresponded to documented warm spells. Together, these results demonstrate that the dataset not only achieves technical quality but also maintains physical interpretability across modalities.

TABLE I

| Dataset | Raw Records | Processed Records | Spatial Resolution | Temporal Coverage | Key Preprocessing Steps |
|---|---|---|---|---|---|
| Sentinel-2 Imagery | ~850 scenes | 600 image patches | 10–20 m | 2017 – 2024 | Cloud masking, band normalization, patch extraction (128×128), data augmentation, manual labeling |
| ITS_LIVE Velocity | 73,590,934 | 11,960,000 | ~120 m | 1988 – 2024 | Daily aggregation, spatial averaging, outlier filtering, velocity magnitude computation, cyclical encoding |
| MODIS LST | 528,704 | 92,736 | 1 km | 2000 – 2023 | QC filtering, interpolation (< 5 days), anomaly computation, normalization, seasonal encoding |
| Integrated Dataset | 54734 | 3 synchronized data streams | Multi-resolution | 2019 – 2023 | Temporal and spatial harmonization, multimodal normalization |

Table I. Overview of the datasets integrated in GLOFNet, showing raw and processed sample sizes, resolutions, temporal spans, and major preprocessing operations.

In summary (Table I), GLOFNet provides a high-quality, multimodal dataset that captures essential cryospheric processes relevant to GLOF hazards. Seasonal cycles, long-term warming trends, spatial heterogeneity, and event precursors are all preserved in a form suitable for deep learning and hazard forecasting.

## V. Data Availability

The GLOFNet dataset introduced in this paper, including all preprocessed Sentinel-2 imagery, ITS_LIVE glacier velocity data, and MODIS LST sequences, is publicly available at: https://drive.google.com/drive/folders/191x2uwFRzgd2CMfqpqdVw0UrT5YZYjHN.

## VI. Conclusion

This study presented GLOFNet, a comprehensive multimodal dataset designed to support research on Glacial Lake Outburst Flood (GLOF) monitoring and prediction. By integrating Sentinel-2 multispectral imagery, NASA ITS_LIVE glacier velocity products, and MODIS Land Surface Temperature (LST) records, GLOFNet captures complementary spatial, kinematic, and thermal indicators of glacier instability. The dataset spans more than three decades (1988–2024), covering the full evolution of Shisper Glacier from quiescence to surge and lake outburst.

Through a rigorous preprocessing and harmonization pipeline, the raw satellite observations were transformed into a high-quality, temporally consistent, and physically interpretable dataset. Exploratory analyses revealed clear seasonal and interannual patterns in glacier velocity and temperature, validating the robustness of the data and its suitability for machine learning applications. The resulting multimodal structure provides a foundation for both empirical studies of cryospheric processes and the development of deep learning frameworks for hazard forecasting.

Despite challenges such as limited GLOF event occurrences, cloud contamination, and resolution mismatches among sensors, GLOFNet establishes a scalable blueprint for data fusion in glacier hazard research. Future extensions will focus on incorporating Synthetic Aperture Radar (SAR) data to mitigate optical limitations, improving spatial downscaling of thermal observations, and expanding coverage across multiple glacier basins to enhance model generalizability.

In summary, GLOFNet represents a first-of-its-kind, openly curated dataset that bridges visual, physical, and thermal dimensions of glacier dynamics, enabling the scientific community to advance toward more accurate, interpretable, and operational GLOF early-warning systems.

Acknowledgment

This research was supported by the resources and guidance provided by my supervisors and institution, whose contributions are gratefully acknowledged.